\documentclass[runningheads]{llncs}
\usepackage[T1]{fontenc}
\usepackage{graphicx}
\usepackage{xspace}
\usepackage{xcolor}
\usepackage{wrapfig} 
\usepackage{booktabs}
\usepackage{multirow}
\usepackage{makecell}
\usepackage{graphicx}
\usepackage{amssymb}

\usepackage[colorlinks=true,
            linkcolor=blue,     % For internal links (e.g., TOC)
            citecolor=red,      % For citation links
            urlcolor=teal,      % For URLs
            breaklinks=true,pagebackref=true]{hyperref}

\begin{document}
\title{PuzzleMate: Benchmarking MLLMs for Egocentric Puzzle Assistance}

\newcommand{\method}{\textsc{PuzzleMate}\xspace}
\author{Avijit Dasgupta\inst{1}\orcidID{0000-0001-5633-1843} \and
Shayon Dasgupta\inst{2}\orcidID{0009-0009-5622-0209} \and
Zakaria Laskar\inst{3}\orcidID{0000-0003-1414-0808} \and
C. V. Jawahar\inst{1}\orcidID{0000-0001-6767-7057} \and
Karteek Alahari\inst{4}\orcidID{0000-0002-1838-5936}}
\authorrunning{A. Dasgupta et al.}
\institute{IIIT Hyderabad, India \and
IIT BHU, India \and
IISER Thiruvananthapuram, India \and
Univ. Grenoble Alpes, Inria, CNRS, Grenoble INP, LJK, France}
\maketitle              % typeset the header of the contribution
\begin{abstract}
Personal AI assistants hold the potential to evolve from digital interfaces into embodied companions capable of guiding users through complex physical activities. For these assistants to become integral to daily life, they must do more than identify objects; they must provide precise, step-by-step instructions that align with a user's real-time progress. While Multimodal Large Language Models (MLLMs) show promise in general visual understanding, their ability to deliver grounded, sequential guidance for fine-grained manipulation tasks remains largely unverified.

In this paper, we choose the jigsaw puzzle as a strategic testbed for this capability. Unlike general object recognition, puzzle solving demands high-precision spatial reasoning, the ability to distinguish between minute geometric variations, and a rigorous adherence to sequential logic.

We investigate this capability through \method, a novel framework focused on jigsaw puzzle solving captured through an egocentric viewpoint. We deploy \method in a user-in-the-loop study to evaluate how well state-of-the-art MLLMs perceive the current puzzle state and generate actionable next-step instructions. Our analysis reveals seven key bottlenecks that limit their effectiveness. Building on these insights, we propose a benchmark that enables systematic evaluation of MLLMs’ reasoning capabilities for puzzle solving. Our findings reveal a substantial performance gap in current models like GPT-5.2 and Gemini-2.5-Pro; while these MLLMs are highly capable, they struggle to navigate the intricate reasoning and sequential logic essential for jigsaw puzzle assistance.

\end{abstract}
%
%
%

% v1

\section{Introduction}

\begin{figure*}
    \centering
    \includegraphics[width=\linewidth]{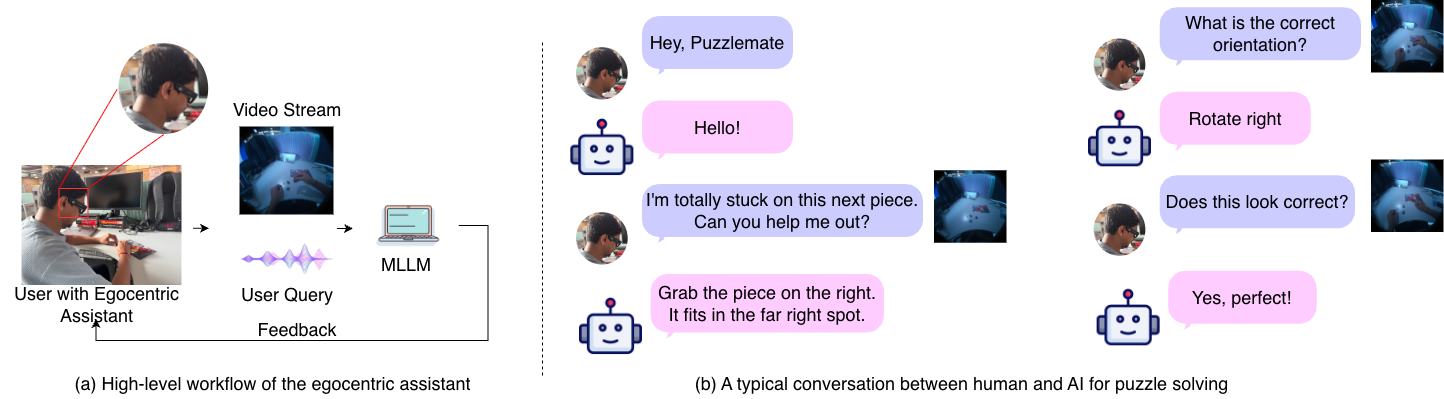}
    \caption{Overview of the proposed egocentric assistive system and interaction. (a) High-level workflow of our egocentric assistant, \textsc{PuzzleMate}, where first-person video from smart glasses and user queries are streamed to an MLLM, hosted on a remote server, to generate grounded responses. (b) An example of a human–AI interaction during puzzle solving, illustrating step-by-step, context-aware assistance through natural language dialogue grounded in egocentric visual input.}
    \label{fig:teaser}
    \vspace{-0.3cm}
\end{figure*}

Personal AI assistants are emerging as a central paradigm in human–AI interaction, aiming to provide intelligent, proactive, and personalized support in everyday activities. Unlike traditional task-oriented systems that respond to explicit commands~\cite{Young2013POMDPBasedSS}, modern personal assistants are expected to continuously perceive their environment, understand user intent, and adapt to individual preferences over time~\cite{pu2025promemassist}. Realizing these capabilities in practice requires assistants to be embedded in the physical world of the user, with access to rich, continuous streams of contextual information. Wearable AI devices, particularly smart glasses~\cite{meta2025rayban,engel2023project}, offer a compelling platform for this purpose by providing an egocentric, always-on view of the user’s environment. Unlike desktop or mobile interfaces that rely on intermittent interaction, such devices enable assistants to observe activities as they unfold, capturing fine-grained visual, temporal, and interaction cues that are critical for understanding user intent in the real world.

Recent advances in MLLMs~\cite{Vasu_2025_CVPR,liu2023llava,liu2024llavanext} provide a strong foundation for realizing personal AI assistants that operate in real-world environments. By jointly modeling vision, language, and other sensory modalities, MLLMs bridge low-level perception with high-level semantic reasoning, enabling unified understanding across heterogeneous and unstructured data streams. Unlike traditional modular pipelines that rely on task-specific components and handcrafted interfaces, MLLMs offer greater flexibility and generalization, allowing a single model to support diverse assistive functions including perception, reasoning, and natural language interaction. These properties are particularly important for wearable, egocentric settings, where assistants must interpret noisy, continuous sensory input and adapt their responses to evolving user context. Building on these capabilities, recent systems such as ExpertAF~\cite{ashutosh2025expertaf} demonstrate that MLLMs can function as task assistants by jointly reasoning over visual observations and language. Through capabilities such as captioning, visual question answering, and multimodal reasoning, MLLMs can interpret egocentric scenes and generate guidance that is grounded in the user’s first-person experience.

Despite these advances, existing benchmarks~\cite{grauman2022ego4d,grauman2024ego,damen2018scaling} provide limited insight into how well these MLLMs~\cite{liu2023llava,team2025gemma,an2025llava} function as egocentric assistants. In realistic assistance scenarios, users engage in iterative interactions with an assistant, asking questions, making decisions, and adjusting their actions based on feedback. Supporting such interactions requires models to maintain task context over time, reason about object identity and spatial relationships, and ground language outputs in the egocentric visual stream. Most benchmarks~\cite{karamolegkou,liu2024right} focus on passive understanding tasks or offline performance measures, and do not explicitly test these requirements. Moreover, evaluating these assistants is further complicated by the characteristics of first-person video. Egocentric footage is often affected by motion blur, rapid viewpoint changes, and partial visibility of task-relevant objects~\cite{li2025challenges}. Objects may appear at small scales or be visible only intermittently, while task context may span long temporal horizons.

To this end, we introduce a user-centered evaluation of MLLM-based assistance in egocentric environments. We build an AI system, \textsc{PuzzleMate}, that streams egocentric video from smart glasses~\cite{engel2023project}, accepts user instructions in natural language, and leverages MLLMs to generate responses grounded in first-person visual input. Using this system, we conduct a user study in which participants interact with the assistant while solving puzzles, enabling analysis of real-world, step-by-step assistance in an egocentric setting (see Fig.~\ref{fig:teaser}). Through this study, we identify recurring failure modes and pain points in current MLLM-based assistants, including challenges arising from visual clutter with distracting objects (e.g., monitors or books) that require accurate localization of the object of interest, as well as illumination variations such as shadows and low-light conditions which are typical of indoor environments. Motivated by these observations, we define seven tasks that capture key challenges in egocentric assistance and construct a benchmark based on these tasks to systematically evaluate both closed-source~\cite{team2023gemini,openai_hello_gpt5_2025} and open-source MLLMs~\cite{liu2023llava,team2025gemma,an2025llava} in real-world egocentric assistance setup, revealing their current limitations and opportunities for improvement.

We adopt puzzle solving as the experimental test bed due to its natural alignment with the core requirements of personal AI assistants. Puzzle solving demands accurate perception of visual elements, understanding of spatial and semantic relationships, and reasoning over intermediate states to plan a solution~\cite{toh2025jumping}. Crucially, effective assistance in this setting often requires step-by-step guidance, where each instruction must be grounded in the user’s current progress and visual context. These characteristics closely mirror real-world assistive scenarios, in which personal AI systems must integrate perception with multi-step reasoning and provide timely, context-aware support. At the same time, puzzles offer a controlled yet sufficiently challenging environment, enabling systematic evaluation of an assistant’s ability to ground instructions in egocentric observations, track task state over time, and adapt its guidance as the interaction unfolds.

In summary, this work presents a systematic study of MLLMs for egocentric assistive AI applications. Our key contributions are as follows:
\begin{itemize}
    \item
We design and implement a real-time AI system that streams egocentric video from smart glasses, accepts natural language user instructions, and leverages MLLMs to generate responses grounded in the user’s first-person visual context.
\item
Using this system, we conduct a controlled user study on egocentric puzzle-solving tasks, enabling systematic analysis of how users interact with MLLM-based assistants in real-world settings and identifying key pain points and failure modes related to perception, temporal reasoning, and instruction grounding.
\item
Motivated by insights from the user study, we define seven representative tasks that capture core challenges in egocentric assistive AI and provide a benchmark .

\item We also evaluate both closed-source and open-source MLLMs on these tasks, providing quantitative and qualitative insights into their current capabilities and limitations.

\end{itemize}

Our study reveals that while models perform well on simple piece identification and orientation tasks, both open- and closed-source MLLMs fail to handle more complex reasoning scenarios, especially when distractor pieces are present.

\section{Related Works}

\noindent \textbf{Wearable Devices for Task Assistance.}
Advancements in lightweight computing have led to the emergence of wearable assistants across both consumer and research domains. LLM-enabled commercial devices such as Ray-Ban smart glasses~\cite{meta2025rayban}, the AI Pin~\cite{humane2025aipin}, and the AI Friend necklace~\cite{friend2025necklace} illustrate growing interest in always-on AI systems. Recent research has demonstrated the potential of wearable and context-aware agents to support users in physical environments by leveraging multimodal sensory inputs and AI models~\cite{ashutosh2025expertaf,huh2025vid2coach}. 
Egocentric vision is an integral component of these wearable assistants, as it provides a first-person visual stream that captures the user’s environment, object interactions, and task context directly from the user’s perspective~\cite{plizzari2024outlook}. This viewpoint enables assistants to localize objects of interest, reason about ongoing activities, and ground step-by-step guidance in what the user actually sees. At the same time, egocentric vision introduces significant challenges, including frequent viewpoint changes, visual clutter, occlusions, and illumination variations, which complicate robust perception and long-term task understanding~\cite{li2025challenges}. Effectively leveraging egocentric visual input therefore remains a key challenge in building reliable task-oriented wearable assistants.

\noindent \textbf{Multi-Modal Large Language Models.} MLLMs~\cite{liu2023llava,team2025gemma,an2025llava} extend large language models with the ability to perceive, reason, and generate across multiple modalities, typically by combining modality-specific encoders with a pre-trained LLM through learned projection layers. A key milestone toward unified multimodal reasoning was CLIP~\cite{radford2021learning}, which demonstrated the effectiveness of large-scale vision–language pretraining in aligning visual and textual representations within a shared embedding space. Building on this foundation, recent MLLMs have achieved strong performance across a wide range of vision–language tasks, including image captioning~\cite{Peng_2025_CVPR}, object detection~\cite{yin2025rod}, segmentation~\cite{lai2024lisa}, and visual question answering~\cite{chen2024lion}. By integrating perception and language understanding within a single, instruction-following framework, MLLMs reduce reliance on highly specialized task-specific architectures and offer a scalable approach to visual reasoning.

\noindent \textbf{MLLMs for Assistive Tasks.} Despite rapid progress, it remains unclear how effectively current MLLMs can support general-purpose personal AI assistants operating in realistic, day-to-day egocentric settings. Existing large-scale egocentric benchmarks such as Ego4D~\cite{grauman2022ego4d} and Ego-Exo4D~\cite{grauman2024ego} have driven advances in tasks like action recognition and object tracking, but they offer limited coverage of practical assistive scenarios that require fine-grained visual reasoning and real-time human-in-the-loop interaction. ExpertAF~\cite{ashutosh2025expertaf} demonstrates how MLLMs can generate actionable coaching feedback from video by combining expert commentary with corrective visual demonstrations. Huh \textit{et al.}~\cite{huh2025vid2coach} introduce Vid2Coach, which transforms how-to videos into wearable, camera-based task assistants that provide step-wise, accessible instructions and context-aware, mixed-initiative feedback through egocentric vision. Verghese \textit{et al.}~\cite{verghese2025user} study MLLM-based assistive agents for multi-step activities using both offline video benchmarks and an egocentric user study, highlighting challenges in grounding long visual histories and supporting replanning during task execution. Dasgupta \textit{et al.}~\cite{dasgupta2026} present one of the earliest evaluations of MLLMs in assistive settings, focusing specifically on tasks tailored for individuals with visual impairments.

% While these works primarily focus on structured procedural tasks such as cooking or physical skills, our work aims to go beyond these activities to study egocentric assistance in a more general, controlled and reasoning-intensive setting. We adopt puzzle solving as a testbed, as it requires fine-grained visual perception, multi-step reasoning over intermediate states, and adaptive, step-by-step guidance—core capabilities expected of general-purpose personal AI assistants.

In contrast, we adopt puzzle solving as a testbed to contrast with prior assistive domains such as cooking, where evaluations are often conducted in relatively constrained and scripted settings. While cooking also requires perception and reasoning, puzzle solving provides a simpler and more controlled environment that isolates general task assistance skills under diverse visual contexts and frequent distractors, with minimal procedural priors. 

%Crucially, each puzzle slot constitutes a distinct data point, continuously changing the visual and spatial context, which makes it particularly well suited for systematically testing fine-grained spatial reasoning and contextual grounding expected of general-purpose personal AI assistants.
\section{PuzzleMate: System Overview}

Fig.~\ref{fig:teaser} provides an overview of our user-study platform, \textsc{PuzzleMate}.
The system is organized as a modular, web-based pipeline implemented using a React frontend and a set of backend services. The design supports real-time interaction, rapid prototyping, and controlled user studies. The core components include: (1) User Interaction Interface, (2) Input Processing, (3) Visual Context Capture, (4) Vision–Language Reasoning, and (5) Response Generation and Logging. This separation of concerns enables independent iteration on interface design, model integration, and evaluation logic while maintaining a coherent end-to-end user experience.

\paragraph{User Interaction Interface.}
The frontend is implemented in React and supports the user study by providing only essential experiment controls and visual feedback. It displays the live egocentric camera stream and exposes basic controls to start, pause, and stop an experimental session. All task-related interaction occurs through spoken queries and assistant responses; the interface does not present instructions or puzzle-specific guidance. This design ensures that the study captures natural, speech-driven interaction with the assistive system while the UI remains lightweight and non-intrusive, serving primarily as a monitoring and session-management tool.

\paragraph{User Query Processing.}
The system continuously monitors audio input and listens for a predefined wake word (“hey PuzzleMate”) to initiate interaction. Upon detecting the wake word, all subsequent speech is captured and transcribed in a streaming manner. The system automatically segments the query by detecting sustained silence; a pause longer than three seconds is treated as the end of the user’s utterance. The resulting transcription is normalized and formatted into a standardized textual representation before being sent to the backend reasoning module. Each query is time-stamped and associated with the current session to ensure accurate alignment with the corresponding egocentric visual context and model response. The interaction loop continues in this manner until a termination phrase (“thank you”) is detected, at which point the system gracefully ends the session.

\paragraph{Visual Context Capture.}
For each interaction, the system captures the relevant visual context of the puzzle directly from the egocentric RGB camera mounted on the Aria glasses~\cite{engel2023project}. Rather than recording the full video stream, the system samples a short temporal window by extracting five consecutive frames at the time a user initiates a query. This multi-frame snapshot provides additional contextual cues about recent user actions and scene dynamics while keeping storage and processing overhead low.

\paragraph{Vision–Language Reasoning.}
The captured image, reference puzzle image, and processed user input are combined into a unified multimodal prompt and sent to the vision–language model. The prompt construction follows a consistent template that embeds both the visual context and the task-specific instruction, enabling fair comparison across models and tasks. The inference module operates independently of the frontend, allowing different MLLMs to be evaluated without modifying the UI or interaction logic. Model outputs are collected along with inference latency for subsequent benchmarking and error analysis.

\paragraph{Response Generation and Logging.}
Model responses are parsed and delivered to the user exclusively through text-to-speech (TTS). The generated response is converted into audio and played back to the user, enabling fully hands-free interaction without any additional visual overlays or cues. All interactions are logged to persistent storage, including the associated egocentric frame sequence ID, transcribed user query, model response text, session identifier, and timing information to support both quantitative evaluation and qualitative analysis of user behavior and failure modes observed during the user study.
\section{Benchmark Setup}

Puzzle solving requires iterations through several interdependent steps: i) given a target slot, determining which candidate piece to select among multiple plausible options, where the number of options is a critical performance factor as increasing candidates reduces the effective spatial resolution available per piece when jointly visible; (ii) given a candidate piece, deciding which slot it should be placed in, again among multiple feasible slots and (iii) reasoning over dependencies among neighboring missing slots to determine an effective placement order. In addition, successful placement requires spatial reasoning to determine the correct orientation or rotation of a selected piece with respect to the target slot. Importantly, these decisions must be communicated and grounded through natural language—specifying slot selection, piece selection, and orientation in a precise and unambiguous manner—which constitutes a core real-world challenge for MLLMs. 

%Based on these interdependent steps, we design seven tasks that isolate and systematically evaluate a MLLM’s ability to perform each of these fundamental operations in realistic puzzle-solving scenarios.

\begin{figure*}[t!]
    \centering
    \includegraphics[width=\linewidth]{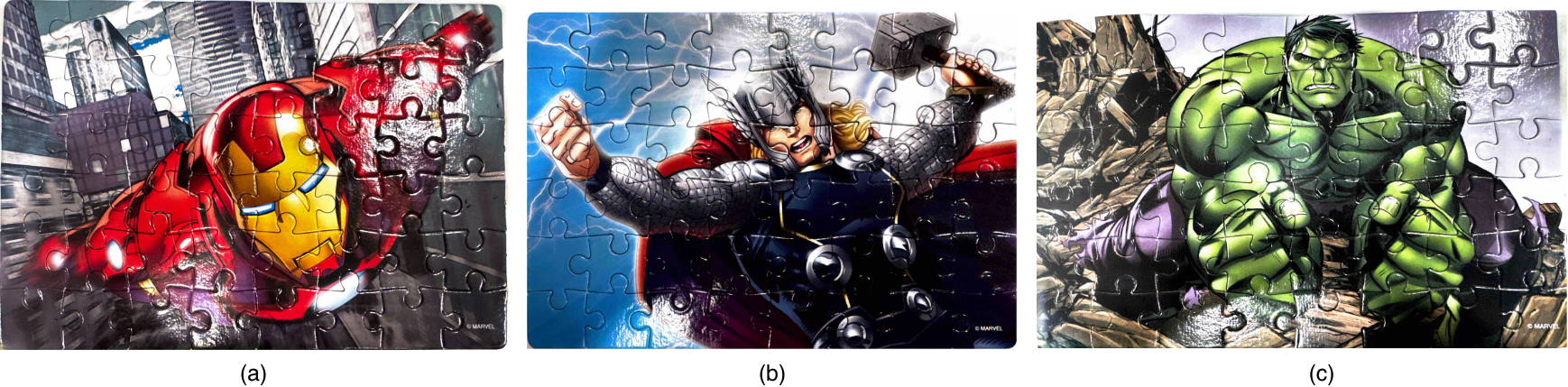}
    \caption{\textbf{Sample egocentric puzzle images used in our benchmark.} Representative 6$\times$8 puzzle instances featuring \textit{Iron Man}, \textit{Hulk}, and \textit{Thor} are used in our evaluation, highlighting both the visual diversity and structural consistency of the benchmark puzzles. \vspace{-0.5cm}}
    \label{fig:puzzle_samples}
\end{figure*}

Based on our assistive AI application, \textsc{PuzzleMate}, we conducted a user study with five participants, each solving three different puzzles with the help of the assistant. Specifically, we explore puzzles of varying difficulty to study the behavior of MLLMs under different levels of spatial complexity. In the user study, we initially used a single medium-difficulty puzzle consisting of 48 pieces (see Fig.~\ref{fig:puzzle_samples}a). However, we observed that current MLLMs were unable to provide meaningful or actionable assistance at this level of complexity, often failing to ground instructions in the visual context. We also experimented with a highly complex puzzle containing over 500 pieces, but found that even human participants were unable to complete it within a reasonable time, making it unsuitable for interactive evaluation. To ensure a fair and informative assessment of assistive interaction, we therefore transitioned to a set of six easy puzzles, each containing 4–6 pieces, for the user study. This setting allowed participants to complete tasks while still exposing common failure modes in instruction grounding and spatial reasoning. In contrast, for the benchmark tasks, we retained higher complexity and used three medium-difficulty puzzles with 48 pieces each (see Fig.~\ref{fig:puzzle_samples}). These puzzles exhibit regular, symmetric geometric structures, which simplify task design and annotation while still enabling systematic evaluation of model performance under more challenging and realistic spatial reasoning conditions. 

% \begin{figure*}[t]
%     \centering
%     \includegraphics[width=\linewidth]{ICPR_2026_LaTeX_Templates/figs/task_1_2.drawio.png}
%     \caption{Overview of representative benchmark tasks used to evaluate egocentric spatial reasoning in MLLMs. Shown here are two of the seven tasks in our benchmark: Task 1 (Piece–Slot Compatibility) evaluates whether a candidate piece fits a given gap; and Task 2 (Piece Orientation Verification) tests the model’s ability to determine whether the piece is correctly rotated for alignment. Together, these tasks probe complementary aspects of spatial reasoning under egocentric visual input.}
%     \label{fig:task_1_2}
% \end{figure*}

\begin{figure*}[h!]
    \centering
    \begin{tabular}{ccc}
    \includegraphics[width=0.45\linewidth]{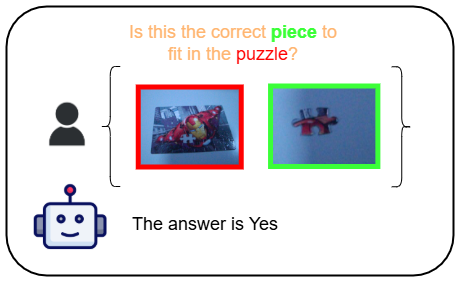} &
    \includegraphics[width=0.45\linewidth]{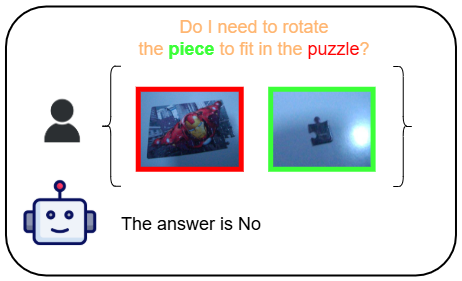} \\
    (a) Task 1 & (b) Task 2 \\
    \includegraphics[width=0.45\linewidth]{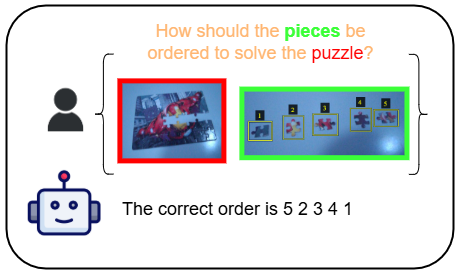} &
    \includegraphics[width=0.45\linewidth]{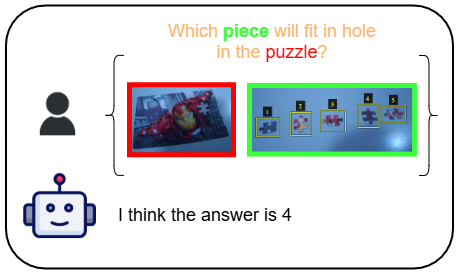} \\
    (a) Task 3 & (b) Task 4 \\
    \includegraphics[width=0.45\linewidth]{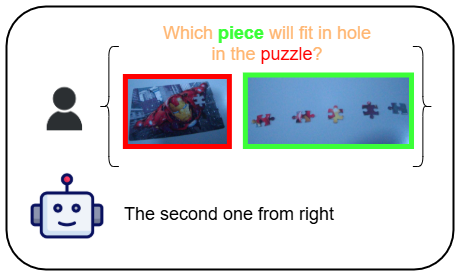} &
    \includegraphics[width=0.45\linewidth]{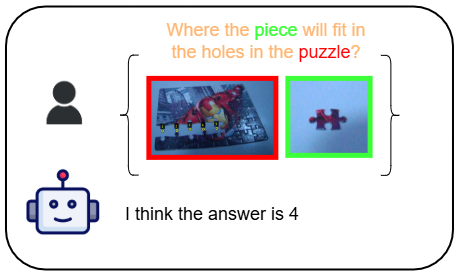} \\
    (a) Task 5 & (b) Task 6 \\
    % \includegraphics[width=\linewidth]{ICPR_2026_LaTeX_Templates/figs/ICPR_TASKS_v1.drawio.png} \\
     % \includegraphics[width=\linewidth]{ICPR_2026_LaTeX_Templates/figs/task_3_4.drawio.png}
     % \\
     % \includegraphics[width=\linewidth]{ICPR_2026_LaTeX_Templates/figs/task_5_6.drawio.png}
    % \label{fig:task_1_2} 
    \end{tabular}
    \caption{Overview of representative benchmark tasks used to evaluate egocentric spatial reasoning in MLLMs. Shown here are two of the seven tasks in our benchmark: Task 1 (Piece–Slot Compatibility) evaluates whether a candidate piece fits a given gap; and Task 2 (Piece Orientation Verification) tests the model’s ability to determine whether the piece is correctly rotated for alignment. Together, these tasks probe complementary aspects of spatial reasoning under egocentric visual input.}
    \label{fig:task_1_6}
\end{figure*}

% Based on our assistive AI application, \textsc{PuzzleMate}, we conducted a user study with five participants, each solving three different puzzles with the help of the assistant. 
% %Our analysis showed that, on average, participants were able to complete $58.4\%$ of the puzzles by following the assistant’s instructions. 
% To better understand the underlying causes of success and failure, we closely examined the interaction transcripts and visual context across all sessions. This qualitative analysis revealed seven recurring pain points in current MLLM-based egocentric assistance, which informed the design of our benchmark tasks. By grounding our task formulation in real user interactions, we ensure that the benchmark reflects practical challenges encountered in realistic assistive scenarios in the context of puzzle solving.

\noindent \textbf{Task 1: Piece–Slot Compatibility.} This task evaluates whether an assistant can determine if a given puzzle piece correctly fits into a missing slot (refer Fig.~\ref{fig:task_1_6} (a)). The input consists of an egocentric view of the puzzle with one missing region and a candidate piece. The model is asked to make a binary decision indicating whether the candidate piece fits the free slot or not. To isolate shape and spatial compatibility, we ensure that the candidate piece is shown in the correct orientation, eliminating errors arising from rotation or alignment. We intentionally restrict the output to a binary prediction to simplify success and failure modes and enable unambiguous evaluation. This task targets a fundamental capability for egocentric assistance, namely accurate visual comparison under real-world conditions, which was frequently observed as a source of failure in the user study.

\noindent \textbf{Task 2: Piece Orientation Verification.} Once the correct puzzle piece is identified in Task 1, determining its correct orientation for placement into the target slot is a natural next step. In Task 2, we evaluate whether the MLLMs can determine if a correctly matching puzzle piece is in the correct orientation for a given missing slot (see Fig.~\ref{fig:task_1_6} (b)). The input consists of an egocentric view of the puzzle with one missing slot and the corresponding piece, which is shown either in a correct or incorrect rotation. The model is required to make a binary decision indicating whether the orientation is correct. By fixing the piece identity and varying only its rotation, this task isolates orientation reasoning from shape compatibility. As in Task 1, we restrict the output to a binary prediction to simplify evaluation and clearly attribute errors to failures in spatial reasoning. This capability is essential for step-by-step assistance, as incorrect orientation guidance was a common source of user confusion observed in the user study.

% Once the correct puzzle piece is identified in Task 1, determining its correct orientation for placement into the target slot naturally follows as Task 2. Together, Tasks 1 and 2 are designed to evaluate a VLM’s ability to reason over contextual information in real-world task-solving scenarios. The context may range from a single isolated piece at early stages of the task to a partially reconstructed puzzle. In this work, we focus on the simplest yet representative setting: an almost fully assembled puzzle with a single missing piece, where the context consists of the near-complete puzzle state. In the subsequent tasks, we move towards a more realistic setting, where multiple pieces are shown jointly at a time

% \begin{figure*}[t]
%     \centering
%     \includegraphics[width=\linewidth]{ICPR_2026_LaTeX_Templates/figs/task_3_4.drawio.png}
%     \caption{Overview of additional benchmark tasks targeting object selection and spatial grounding in egocentric assistance. Task 3 (Piece Sequence Ordering) requires predicting the correct assembly order of multiple pieces to complete a contiguous section of the puzzle. Task 4 (Piece Selection from Multiple Candidates) evaluates the model’s ability to identify the correct piece for a missing slot from several options. These tasks progressively increase the difficulty of visual grounding and spatial correspondence.}
%     \label{fig:task_3_4}
% \end{figure*}
\noindent \textbf{Task 3: Piece Sequence Ordering.} As shown in Fig.~\ref{fig:task_1_6} (c), we evaluate whether an assistant can infer the correct sequence of puzzle pieces for a contiguous set of missing regions. The input consists of an egocentric view of the puzzle with five consecutive missing pieces and five corresponding candidate pieces provided in an unordered manner. The candidate pieces are assumed to be shown in the correct orientation, allowing the task to focus solely on sequence reasoning rather than rotation. To reduce ambiguity in referring to pieces, we draw bounding boxes around each candidate piece and assign them unique identifiers, and the model is required to predict the correct ordering of these identifiers. This task evaluates the model’s ability to reason over spatial continuity across multiple positions and maintain consistency over several steps.

\noindent \textbf{Task 4: Piece Selection from Multiple Candidates.} In this task, we evaluate whether an assistant can identify the correct puzzle piece for a given missing slot from multiple candidates (see Fig.~\ref{fig:task_1_6} (d)). The input consists of an egocentric view of the puzzle with a single missing region and five candidate pieces, each highlighted with a bounding box and assigned a unique identifier. This task extends Task 1 by moving beyond binary compatibility judgment to explicit piece selection, requiring the model to choose the correct candidate from a set of alternatives. To simplify evaluation and avoid ambiguity, the model’s response is restricted to selecting one of the numbered bounding boxes. This formulation better reflects realistic assistive scenarios, where an assistant must not only assess fit but also identify the correct piece among the candidate pieces.

\noindent \textbf{Task 5: Piece Selection with Natural Language Reference.}
This task follows the same setup as Task 4, where the assistant must identify the correct puzzle piece for a given missing slot from five candidates (refer Fig.~\ref{fig:task_1_6} (e)). However, unlike Task 4, the candidate pieces are not annotated with bounding boxes having numerical identifiers. Instead, the model is required to refer to the correct piece using natural language. This formulation introduces additional challenges in visual grounding and referring expression generation, as the assistant must accurately describe the target piece based on its visual attributes and spatial context. By removing explicit annotations, this task more closely resembles real-world assistive scenarios, where users rely on natural language descriptions rather than pre-defined visual markers. Failures in this setting were common during the user study, often resulting in ambiguous or unhelpful instructions that were difficult for users to follow.

% \begin{figure*}[t]
%     \centering
%     \includegraphics[width=\linewidth]{ICPR_2026_LaTeX_Templates/figs/task_5_6.drawio.png}
%     \caption{Overview of representative benchmark tasks used to evaluate egocentric spatial reasoning in MLLMs. Task 5 (Piece Selection with Natural Language Reference) extends this setting by requiring the model to refer to the correct piece using natural language rather than explicit identifiers; and Task 6 (Piece-to-Location Matching) tests whether the model can determine the correct placement location for a given piece among multiple missing regions. These tasks progressively increase the difficulty of visual grounding and spatial correspondence.}
%     \label{fig:task_5_6}
% \end{figure*}

\noindent \textbf{Task 6: Piece-to-Location Matching.}
In this task, we evaluate whether an assistant can identify the correct location for a given puzzle piece when multiple regions in the puzzle are missing (refer Fig.~\ref{fig:task_1_6} (f)). 
The input consists of an egocentric view of the puzzle with several missing slots, each labeled with a unique identifier, along with a single candidate piece. 
\begin{wrapfigure}{r}{0.35\textwidth}
  \centering
  \includegraphics[width=\linewidth]{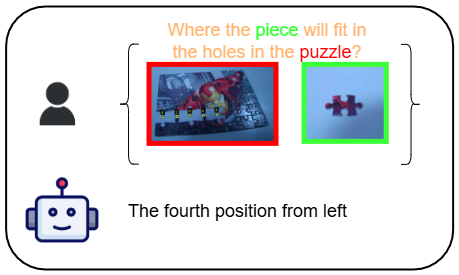}
  \caption{\textbf{Task 7} (Piece-to-Location Matching with Natural Language Reference). The model identifies the correct placement location for a given puzzle piece among multiple missing regions using natural language descriptions.}
  \label{fig:task_7}
\end{wrapfigure} 
The model is required to predict which labeled location the piece belongs to. By reversing the selection direction compared to Task 4, this task isolates the model’s ability to reason about global puzzle context and spatial correspondence across multiple candidate locations. This capability is essential for effective egocentric assistance, as users often need help deciding where a particular piece should be placed among several possible options.

\noindent \textbf{Task 7: Piece-to-Location Matching with Natural Language Reference.} This task builds on Task 6 by removing explicit location identifiers (refer Fig.~\ref{fig:task_7}). The input consists of an egocentric view of the
puzzle with multiple missing regions and a single candidate piece. Instead of selecting a labeled location, the model is required to describe, in natural language, where the piece should be placed. This formulation introduces additional challenges in spatial description and visual grounding, as the assistant must refer to the correct location using relative position, nearby visual cues, or contextual landmarks within the scene.
By requiring natural language output, this task better reflects real-world assistive interactions, where precise yet unambiguous descriptions are necessary for effective guidance. Such referring expressions were a frequent source of failure in the user study.
A table summarizing the dataset size for each tasks is provided in the supplementary.

\section{Evaluation and Analysis}

We evaluate both state-of-the-art closed-source MLLMs, such as Gemini-2.5-Pro~\cite{team2023gemini} and GPT-5.2~\cite{openai_hello_gpt5_2025}, as well as open-source models including LLaVA-1.5~\cite{liu2023llava}, Gemma~\cite{team2025gemma}, and LLaVA-OneVision-1.5~\cite{an2025llava}, spanning parameter scales from 4B to 27B.
We adopt task-specific evaluation strategies aligned with the nature of each benchmark task. For Tasks 1, 2, 4, and 6, which require discrete decisions or selections, we use accuracy as the evaluation metric. These tasks have unambiguous ground-truth labels, allowing straightforward measurement of success or failure.
\begin{table*}[t]
\centering
\caption{Performance comparison of MLLMs across the seven benchmark tasks. \textcolor{green}{$\blacktriangle$} denotes percentage improvement when using Aria relative to GoPro, while \textcolor{red}{$\blacktriangledown$} denotes performance degradation.}
\label{tab:benchmark_results}
\renewcommand{\arraystretch}{1.2}
\setlength{\tabcolsep}{6pt}

\resizebox{\textwidth}{!}{%
\begin{tabular}{llcccccccc}
\toprule
\multirow{2}{*}{\textbf{Camera}} 
& \multirow{2}{*}{\textbf{Model}} 
& \textbf{Task 1} 
& \textbf{Task 2} 
& \multicolumn{2}{c}{\textbf{Task 3}} 
& \textbf{Task 4} 
& \textbf{Task 5} 
& \textbf{Task 6} 
& \textbf{Task 7} \\
\cmidrule(lr){5-6}
& 
& (Acc. $\uparrow$) 
& (Acc. $\uparrow$) 
& (Kendall's Tau $\uparrow$) 
& (Edit Dist. $\downarrow$) 
& (Acc. $\uparrow$) 
& (Acc. $\uparrow$) 
& (Acc. $\uparrow$) 
& (Acc. $\uparrow$) \\
\midrule

\multirow{8}{*}{\textbf{GoPro\footnote{\url{https://gppro.in/product/gopro-hero10-black/}} }}
& Gemini-2.5-Pro~\cite{team2023gemini}      & 0.71 & 0.92 & 0.19 & 2.55 & 0.44 & 0.54 & 0.50 & 0.53 \\
& GPT-5.2~\cite{openai_hello_gpt5_2025}            & 0.69 & 0.93 & 0.35 & 2.33 & 0.56 & 0.56 & 0.54 & 0.58 \\
& LLaVA-1.5-7B~\cite{liu2023llava}       & 0.54 & 0.50 & 0.28 & 2.00 & 0.26 & 0.25 & 0.33 & 0.33 \\
& LLaVA-1.5-13B~\cite{liu2023llava}         & 0.58 & 0.93 & 0.28 & 2.03 & 0.27 & 0.27 & 0.34 & 0.35 \\
& Gemma-3-4B~\cite{team2025gemma}          & 0.57 & 0.54 & 0.14 & 3.10 & 0.34 & 0.42 & 0.30 & 0.35 \\
& Gemma-3-12B~\cite{team2025gemma}         & 0.63 & 0.91 & 0.29 & 3.20 & 0.30 & 0.44 & 0.37 & 0.40 \\
& Gemma-3-27B~\cite{team2025gemma}         & 0.64 & 0.90 & 0.34 & 2.28 & 0.30 & 0.46 & 0.40 & 0.42 \\
& LLaVA-OneVision-1.5-8B~\cite{an2025llava} & 0.56 & 0.50 & 0.28 & 1.98 & 0.29 & 0.31 & 0.38 & 0.37 \\

\midrule

\multirow{8}{*}{\textbf{Aria}}
& Gemini-2.5-Pro~\cite{team2023gemini}      & 0.68 \textcolor{red}{$\blacktriangledown$ 3.0\%}   & 0.91 \textcolor{red}{$\blacktriangledown$ 1.0\%} & 0.08 \textcolor{red}{$\blacktriangledown$ 57.89.0\%} & 3.00 \textcolor{red}{$\blacktriangledown$ 17.64\%} & 0.41 \textcolor{red}{$\blacktriangledown$ 3.0\%} & 0.48 \textcolor{red}{$\blacktriangledown$ 6.0\%} & 0.42 \textcolor{red}{$\blacktriangledown$ 8.0\%} & 0.48 \textcolor{red}{$\blacktriangledown$ 5.0\%} \\

& GPT-5.2~\cite{openai_hello_gpt5_2025}            & 0.66 \textcolor{red}{$\blacktriangledown$ 3.0\%}  & 0.92 \textcolor{red}{$\blacktriangledown$ 1.0\%} & 0.29 \textcolor{red}{$\blacktriangledown$ 17.14\%}  & 2.01 \textcolor{green}{$\blacktriangle$ 13.73\%}  & 0.50 \textcolor{red}{$\blacktriangledown$ 6.0\%} & 0.52 \textcolor{red}{$\blacktriangledown$ 4.0\%} & 0.46 \textcolor{red}{$\blacktriangledown$ 8.0\%} & 0.52 \textcolor{red}{$\blacktriangledown$ 6.0\%} \\

& LLaVA-1.5-7B~\cite{liu2023llava}        & 0.53 \textcolor{red}{$\blacktriangledown$ 1.0\%} & 0.50 \textcolor{red}{$\blacktriangledown$ 0.0\%} & 0.28 \textcolor{red}{$\blacktriangledown$ 0.0\%} & 2.00 \textcolor{red}{$\blacktriangledown$ 0.0\%} & 0.25 \textcolor{red}{$\blacktriangledown$ 1.0\%} & 0.25 \textcolor{red}{$\blacktriangledown$ 0.0\%} & 0.33 \textcolor{red}{$\blacktriangledown$ 0.0\%} & 0.33 \textcolor{red}{$\blacktriangledown$ 0.0\%}\\

& LLaVA-1.5-13B~\cite{liu2023llava}         & 0.56 \textcolor{red}{$\blacktriangledown$ 2.0\%} & 0.90 \textcolor{red}{$\blacktriangledown$ 3.0\%} & 0.26 \textcolor{red}{$\blacktriangledown$ 7.14\%} & 2.03 \textcolor{red}{$\blacktriangledown$ 0.0\%} & 0.27 \textcolor{red}{$\blacktriangledown$ 0.0\%} & 0.27 \textcolor{red}{$\blacktriangledown$ 0.0\%} & 0.33 \textcolor{red}{$\blacktriangledown$ 1.0\%} & 0.33 \textcolor{red}{$\blacktriangledown$ 2.0\%}\\

& Gemma-3-4B~\cite{team2025gemma}          & 0.54 \textcolor{red}{$\blacktriangledown$ 3.0\%} & 0.50 \textcolor{red}{$\blacktriangledown$ 4.0\%} & 0.23 \textcolor{green}{$\blacktriangle$ 64.28\%} & 3.35 \textcolor{red}{$\blacktriangledown$ 8.06\%} & 0.34 \textcolor{red}{$\blacktriangledown$ 0.0\%} & 0.42 \textcolor{red}{$\blacktriangledown$ 0.0\%} & 0.30 \textcolor{red}{$\blacktriangledown$ 0.0\%} & 0.28 \textcolor{red}{$\blacktriangledown$ 7.0\%}\\

& Gemma-3-12B~\cite{team2025gemma}         & 0.59 \textcolor{red}{$\blacktriangledown$ 4.0\%} & 0.91 \textcolor{red}{$\blacktriangledown$ 0.0\%} & 0.25 \textcolor{red}{$\blacktriangledown$ 13.79\%} & 3.60 \textcolor{red}{$\blacktriangledown$ 12.5\%} & 0.30 \textcolor{red}{$\blacktriangledown$ 0.0\%} & 0.43 \textcolor{red}{$\blacktriangledown$ 1.0\%} & 0.35 \textcolor{red}{$\blacktriangledown$ 2.0\%} & 0.37 \textcolor{red}{$\blacktriangledown$ 3.0\%} \\

& Gemma-3-27B~\cite{team2025gemma}         & 0.63 \textcolor{red}{$\blacktriangledown$ 1.0\%} & 0.89 \textcolor{red}{$\blacktriangledown$ 1.0\%} & 0.18 \textcolor{red}{$\blacktriangledown$ 47.05\%} & 3.10 \textcolor{red}{$\blacktriangledown$ 35.96\%} & 0.28 \textcolor{red}{$\blacktriangledown$ 2.0\%}& 0.45 \textcolor{red}{$\blacktriangledown$ 1.0\%} & 0.40 \textcolor{red}{$\blacktriangledown$ 0.0\%} & 0.41 \textcolor{red}{$\blacktriangledown$ 1.0\%}\\

& LLaVA-OneVision-1.5-8B~\cite{an2025llava} & 0.54 \textcolor{red}{$\blacktriangledown$ 2.0\%} & 0.50 \textcolor{red}{$\blacktriangledown$ 0.0\%} & 0.28 \textcolor{red}{$\blacktriangledown$ 0.0\%} & 2.00 \textcolor{red}{$\blacktriangledown$ 1.01\%} & 0.27 \textcolor{red}{$\blacktriangledown$ 2.0\%} & 0.30 \textcolor{red}{$\blacktriangledown$ 1.0\%} & 0.34 \textcolor{red}{$\blacktriangledown$ 4.0\%} & 0.35 \textcolor{red}{$\blacktriangledown$ 2.0\%}\\

\bottomrule
\end{tabular}
}
\vspace{-0.3cm}
\end{table*}

Task 3 requires predicting an ordered sequence of puzzle pieces. To capture both ordering correctness and partial alignment with the ground truth, we evaluate this task using Kendall’s $\tau$ to measure rank correlation and Edit distance to quantify sequence-level discrepancies. Together, these metrics provide a comprehensive assessment of sequence reasoning performance.
Tasks 5 and 7 involve generating texture or spatial references in natural language. Since the correctness of these outputs depends on clarity and interpretability rather than exact string matching, we evaluate them through human verification. Annotators judge whether the response is clear, unambiguous, and correctly conveys the intended spatial information, and accuracy is computed based on whether the instruction is deemed correct by human evaluators. The prompts for each task is provided in the supplementary.

\noindent\textbf{Quantitative Evaluation.} We summarize the overall quantitative results for each tasks in Tab.~\ref{tab:benchmark_results}. Overall, models perform substantially better on Tasks 1 and 2 than on Task 3, highlighting a clear gap between local perceptual reasoning and global structural reasoning. Across models, Task 1 achieves accuracies in the range of approximately 0.53–0.68, while Task 2 reaches consistently higher accuracies of around $\sim0.90$ for most models. In contrast, performance on Task 3 is markedly lower, with Kendall’s Tau values typically below 0.35 and relatively high edit distances. This sharp drop indicates that while MLLMs can reliably answer localized visual queries, they struggle when required to jointly reason over multiple pieces and infer a coherent global ordering, underscoring limitations in long-range spatial reasoning and structured state modeling. Across all metrics, closed-source models~\cite{openai_hello_gpt5_2025,team2023gemini} achieve higher performance than open-source models~\cite{team2025gemma,an2025llava,liu2023llava}, pointing to a performance gap in spatial reasoning and multimodal grounding.

\begin{figure*}[h!]
    \centering
    \begin{tabular}{cc}
    \includegraphics[width=0.45\linewidth]{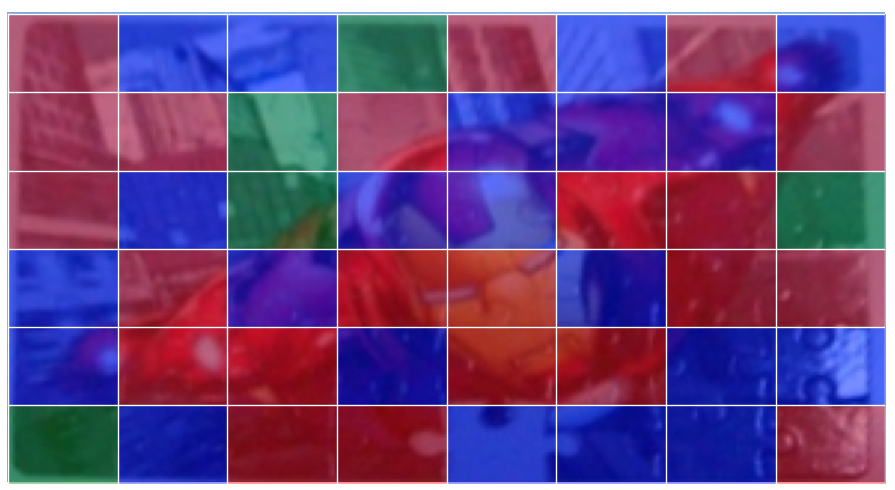} &
    \includegraphics[width=0.49\linewidth, height=0.245\linewidth]{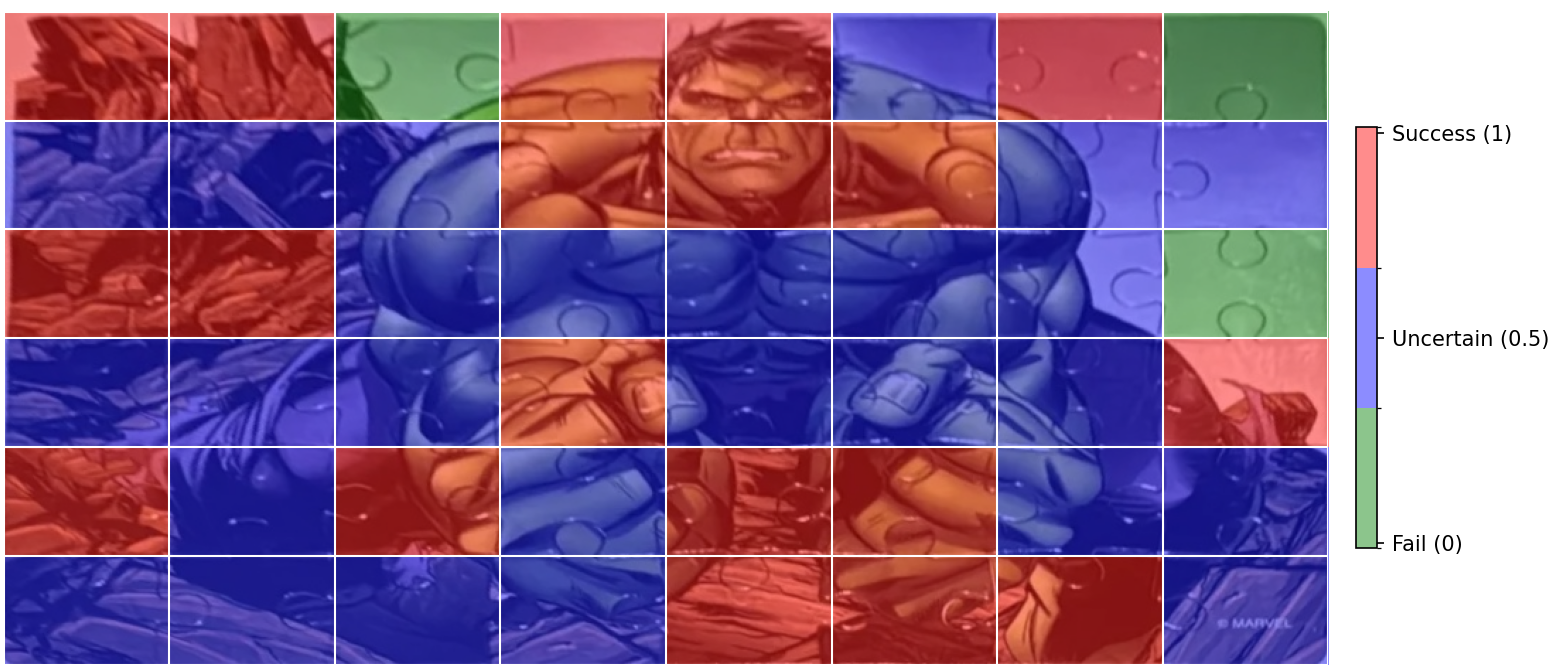} \\
    (a) Ironman & (b) Hulk

    \end{tabular}
    \caption{\textbf{Per-piece prediction confidence visualization across the puzzle.} The heatmap, overlaid on the puzzle image, visualizes the Gemini-2.5-Pro~\cite{team2023gemini} model’s average classification accuracy for Task 1. Each cell corresponds to a puzzle region and reports the mean outcome across the two candidate pieces: \textcolor{red}{1} indicates both pieces are correctly classified, \textcolor{green}{0} indicates both are misclassified, and \textcolor{blue}{0.5} indicates one correct and one incorrect prediction.\vspace{-0.5cm}}
    \label{fig:task_1_heatmap}
\end{figure*}

% \begin{figure*}[h!]
%     \centering
%     \includegraphics[width=\linewidth]{ICPR_2026_LaTeX_Templates/figs/heatmap.drawio.png}
%     \caption{\textbf{Per-piece prediction confidence visualization across the puzzle.} The heatmap shows the model’s predicted confidence for individual puzzle pieces, overlaid on the input image. The model exhibits slightly higher uncertainty toward the puzzle boundaries and edge regions. This trend is plausibly explained by the reduced availability of contextual cues at the periphery, where fewer adjacent pieces provide shape, texture, and spatial constraints. Central regions, by contrast, benefit from richer local context, leading to more stable predictions. The observed effect is mild, indicating that edge-related uncertainty contributes only marginally to overall performance variation.}
%     \label{fig:heatmap}
% \end{figure*}

% \input{ICPR_2026_LaTeX_Templates/sec/table/aria}
% \input{ICPR_2026_LaTeX_Templates/sec/table/gopro}

\noindent\textbf{Qualitative Evaluation.} We present the per-piece prediction performance of Gemini-Pro-2.5~\cite{team2023gemini} in Task 1 across the puzzle in Fig.~\ref{fig:task_1_heatmap}. The model exhibits slightly higher uncertainty toward the puzzle boundaries and edge regions. This trend is plausibly explained by the reduced availability of contextual cues at the periphery, where fewer adjacent pieces provide shape, texture, and spatial constraints. Central regions, by contrast, benefit from richer local context, leading to more stable predictions. The observed effect is mild, indicating that edge-related uncertainty contributes only marginally to overall performance variation. 

\begin{figure*}[t!]
    \centering
    \includegraphics[width=0.8\linewidth]{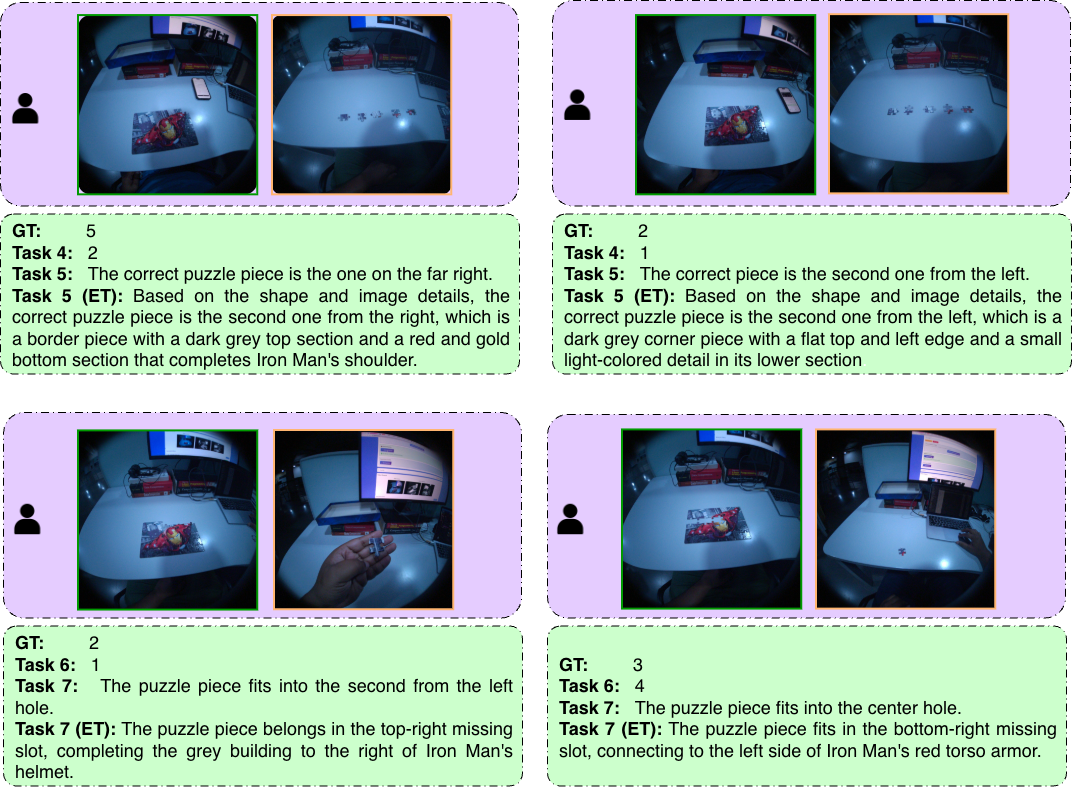}
    \caption{\textbf{Qualitative comparison of model behavior across task formulations}. The top row contrasts Task 4 and Task 5, and the bottom row contrasts Task 6 and Task 7. Constrained, identifier-based outputs in Task 4 and Task 6 often lead to incorrect predictions, whereas allowing concise natural language responses in Task 5 and Task 7 with a limited token budget (35 tokens) yields clearer and more accurate assistance. When the token limit is removed (ET), models generate verbose and ambiguous responses, reducing usability. Overall, the figure illustrates how output modality and token constraints critically influence MLLM performance in egocentric assistive settings. \vspace{-0.7cm}}
    \label{fig:qualitative_4_5_6_7}
\end{figure*}

For Task~4, models are required to produce constrained, identifier-based outputs, and we observe that this formulation frequently leads to incorrect predictions, even when the relevant visual evidence is present (see Fig.~\ref{fig:qualitative_4_5_6_7}). In contrast, Task~5 poses the same underlying reasoning problem but allows the model to respond in natural language under a limited token budget, resulting in more accurate and more interpretable answers and a modest but consistent improvement in performance. These gains are also reflected quantitatively in Tab.~\ref{tab:benchmark_results}. A similar trend is observed for Task~6 versus Task~7: enforcing explicit, structured outputs in Task~6 often leads to failures in correct piece selection, whereas permitting concise natural language explanations in Task~7 enables the model to better articulate its reasoning and identify correct spatial references, again yielding improved quantitative results (Tab.~\ref{tab:benchmark_results}). Overall, when constrained to operate over bounding boxes and discrete identifiers, models struggle to reliably associate visual regions with their semantic roles; allowing natural language responses provides greater flexibility to exploit additional cues such as color, texture, and relative spatial context, leading to more accurate and actionable decisions.

\noindent\textbf{Effect of Extended Tokens.} We further observe that increasing the allowed output tokens (from 35 tokens to default 65,536 tokens in Gemini-Pro-2.5~\cite{team2023gemini}) in Task~5 and Task~7 often degrades performance rather than improving it (Fig.~\ref{fig:qualitative_4_5_6_7}). When extended responses are permitted, models tend to generate verbose and weakly grounded explanations that introduce speculative details not supported by the visual evidence. In spatially precise tasks, such verbosity amplifies early perceptual or association errors. For instance, a model may correctly identify a piece’s location but then hallucinate irrelevant attributes (e.g., referring to Iron Man’s helmet), which conflicts with accurate positional cues and confuses the user. As a result, extended-token (ET) settings dilute decisive visual signals with extraneous text, reducing the reliability and usability of the generated assistance. We provide further qualitative results in the supplementary.

\noindent\textbf{Effect of Camera Quality.} Aria~\cite{engel2023project} images are inherently low resolution and noisy, which is expected for personal AI assistants operating on low-powered, wearable devices. To disentangle the impact of visual quality from reasoning capability, we additionally collected the same set of task inputs using a GoPro  camera, providing higher-resolution and cleaner visual observations. As shown in Table~\ref{tab:benchmark_results}, models consistently achieve higher performance on the GoPro-based evaluation across tasks, indicating that improved visual fidelity directly benefits downstream reasoning and decision-making. This comparison highlights that a significant portion of current failure cases stems from perceptual limitations rather than purely reasoning deficits, while also emphasizing the importance of developing models that are robust to the realistic sensing constraints imposed by wearable, on-device AI systems.
\section{Conclusion}

% In this work, we introduced an egocentric, interactive puzzle-solving benchmark and user-study platform that systematically probes the reasoning capabilities of modern MLLMs under varying task formulations, output constraints, and spatial complexities. Our results show that while current models demonstrate strong visual perception and language generation in isolation, they struggle to consistently translate visual observations into precise, actionable assistance, especially for tasks requiring integration of local visual cues with global spatial structure. We identify key limitations in visual–semantic association, sensitivity to output modality, and robustness to extended reasoning, where additional generation often degrades rather than improves performance. At the same time, our findings point to promising directions: concise natural language responses reliably outperform rigid, identifier-based outputs, suggesting that flexible linguistic grounding can partially mitigate current shortcomings. Looking forward, effective egocentric assistants will require tighter coupling between perception, spatial memory, and controllable reasoning, as well as mechanisms to regulate verbosity and prevent hallucinated elaboration. Addressing these challenges is critical for advancing from passive visual understanding toward dependable, real-time assistive systems capable of operating in human-centered, spatially grounded environments.

In this work, we present an egocentric, interactive puzzle-solving benchmark and user-study platform to systematically evaluate the spatial reasoning capabilities of modern MLLMs. Our results show that although current models exhibit strong visual perception and language generation, they often fail to convert visual observations into precise and actionable assistance, particularly when reasoning requires aligning local visual cues with global spatial structure. We identify key limitations in visual semantic association, sensitivity to output modality, and robustness to extended generation, where increased verbosity frequently degrades performance. Notably, concise natural language responses consistently outperform rigid identifier-based outputs, suggesting that flexible linguistic grounding can partially mitigate these challenges. Looking ahead, reliable egocentric assistants will require tighter integration of perception, spatial reasoning, and controlled task guidance generation to enable dependable real-time human-centered assistance.

\section*{Acknowledgements}
This work was supported by MeitY, Government of India, through the NLTM-Bhashini project. Karteek Alahari was supported in part by the Institute of Information \& Communications Technology Planning \& Evaluation (IITP) grant funded by the Korean Government (MSIT) (No. RS-2024-00457882, National AI Research Lab Project).
% \input{ICPR_2026_LaTeX_Templates/sec/results}

\bibliographystyle{splncs04}
\bibliography{mybibliography}

@inproceedings{huh2025vid2coach,
  title={Vid2Coach: Transforming How-To Videos into Task Assistants},
  author={Huh, Mina and Xue, Zihui and Das, Ujjaini and Ashutosh, Kumar and Grauman, Kristen and Pavel, Amy},
  booktitle={UIST},
  year={2025}
}

@inproceedings{verghese2025user,
  title={User-in-the-loop Evaluation of Multimodal LLMs for Activity Assistance},
  author={Verghese, Mrinal and Chen, Brian and Eghbalzadeh, Hamid and Nagarajan, Tushar and Desai, Ruta},
  booktitle={WACV},
  year={2025}
}

@article{Young2013POMDPBasedSS,
  title={POMDP-Based Statistical Spoken Dialog Systems: A Review},
  author={Steve J. Young and Milica Gasic and Blaise Thomson and J. Williams},
  journal={Proceedings of the IEEE},
  year={2013}
}

@inproceedings{pu2025promemassist,
  title={ProMemAssist: Exploring Timely Proactive Assistance Through Working Memory Modeling in Multi-Modal Wearable Devices},
  author={Pu, Kevin and Zhang, Ting and Sendhilnathan, Naveen and Freitag, Sebastian and Sodhi, Raj and Jonker, Tanya R},
  booktitle={UIST},
  year={2025}
}

@misc{meta2025rayban,
  title        = {Ray-Ban Meta Smart Glasses},
  author       = {{Meta}},
  year         = {2025},
  howpublished = {\url{https://www.ray-ban.com/rayban-meta-ai-glasses}},
  note         = {Accessed April 2025}
}

@misc{humane2025aipin,
  title        = {Humane {AI} Pin: See the World, Not Your Screen},
  author       = {{Humane}},
  year         = {2025},
  howpublished = {\url{https://humane.com/}},
  note         = {Accessed April 2025}
}

@misc{friend2025necklace,
  title        = {Friend: {AI} Necklace},
  author       = {{Friend}},
  year         = {2025},
  howpublished = {\url{https://www.friend.com/wearable/index.html}},
  note         = {Accessed April 2025}
}

@article{engel2023project,
  title={Project aria: A new tool for egocentric multi-modal ai research},
  author={Engel, Jakob and Somasundaram, Kiran and Goesele, Michael and Sun, Albert and Gamino, Alexander and Turner, Andrew and Talattof, Arjang and Yuan, Arnie and Souti, Bilal and Meredith, Brighid and others},
  journal={arXiv preprint arXiv:2308.13561},
  year={2023}
}

@InProceedings{Vasu_2025_CVPR,
    author    = {Vasu, Pavan Kumar Anasosalu and Faghri, Fartash and Li, Chun-Liang and Koc, Cem and True, Nate and Antony, Albert and Santhanam, Gokula and Gabriel, James and Grasch, Peter and Tuzel, Oncel and Pouransari, Hadi},
    title     = {FastVLM: Efficient Vision Encoding for Vision Language Models},
    booktitle = {CVPR},
    year      = {2025}
}

@misc{liu2024llavanext,
    title={LLaVA-NeXT: Improved reasoning, OCR, and world knowledge},
    url={https://llava-vl.github.io/blog/2024-01-30-llava-next/},
    author={Liu, Haotian and Li, Chunyuan and Li, Yuheng and Li, Bo and Zhang, Yuanhan and Shen, Sheng and Lee, Yong Jae},
    month={January},
    year={2024}
}

@InProceedings{liu2023llava,
      title={Visual Instruction Tuning}, 
      author={Liu, Haotian and Li, Chunyuan and Wu, Qingyang and Lee, Yong Jae},
      booktitle={NeurIPS},
      year={2023},
}

@inproceedings{ashutosh2025expertaf,
  title={ExpertAF: Expert actionable feedback from video},
  author={Ashutosh, Kumar and Nagarajan, Tushar and Pavlakos, Georgios and Kitani, Kris and Grauman, Kristen},
  booktitle={CVPR},
  year={2025}
}

@inproceedings{grauman2022ego4d,
  title={Ego4d: Around the world in 3,000 hours of egocentric video},
  author={Grauman, Kristen and Westbury, Andrew and Byrne, Eugene and Chavis, Zachary and Furnari, Antonino and Girdhar, Rohit and Hamburger, Jackson and Jiang, Hao and Liu, Miao and Liu, Xingyu and others},
  booktitle={CVPR},
  year={2022}
}

@inproceedings{grauman2024ego,
  title={Ego-exo4d: Understanding skilled human activity from first-and third-person perspectives},
  author={Grauman, Kristen and Westbury, Andrew and Torresani, Lorenzo and Kitani, Kris and Malik, Jitendra and Afouras, Triantafyllos and Ashutosh, Kumar and Baiyya, Vijay and Bansal, Siddhant and Boote, Bikram and others},
  booktitle={CVPR},
  year={2024}
}

@inproceedings{damen2018scaling,
  title={Scaling egocentric vision: The epic-kitchens dataset},
  author={Damen, Dima and Doughty, Hazel and Farinella, Giovanni Maria and Fidler, Sanja and Furnari, Antonino and Kazakos, Evangelos and Moltisanti, Davide and Munro, Jonathan and Perrett, Toby and Price, Will and others},
  booktitle={ECCV},
  year={2018}
}

@article{li2025challenges,
  title={Challenges and Trends in Egocentric Vision: A Survey},
  author={Li, Xiang and Qiu, Heqian and Wang, Lanxiao and Zhang, Hanwen and Qi, Chenghao and Han, Linfeng and Xiong, Huiyu and Li, Hongliang},
  journal={arXiv preprint arXiv:2503.15275},
  year={2025}
}

@article{toh2025jumping,
  title={The Jumping Reasoning Curve? Tracking the Evolution of Reasoning Performance in GPT-[n] and o-[n] Models on Multimodal Puzzles},
  author={Toh, Vernon YH and Chia, Yew Ken and Ghosal, Deepanway and Poria, Soujanya},
  journal={arXiv preprint arXiv:2502.01081},
  year={2025}
}

@inproceedings{karamolegkou,
    title = "Evaluating Multimodal Language Models as Visual Assistants for Visually Impaired Users",
    author = "Karamolegkou, Antonia  and
      Nikandrou, Malvina  and
      Pantazopoulos, Georgios  and
      Sanchez Villegas, Danae  and
      Rust, Phillip  and
      Dhar, Ruchira  and
      Hershcovich, Daniel  and
      S{\o}gaard, Anders",
    year = "2025"
}

@article{liu2024right,
  title={Right this way: Can VLMs Guide Us to See More to Answer Questions?},
  author={Liu, Li and Yang, Diji and Zhong, Sijia and Tholeti, Kalyana Suma Sree and Ding, Lei and Zhang, Yi and Gilpin, Leilani},
  journal={NeurIPS},
  year={2024}
}

@article{team2025gemma,
  title={Gemma 3 technical report},
  author={Team, Gemma and Kamath, Aishwarya and Ferret, Johan and Pathak, Shreya and Vieillard, Nino and Merhej, Ramona and Perrin, Sarah and Matejovicova, Tatiana and Ram{\'e}, Alexandre and Rivi{\`e}re, Morgane and others},
  journal={arXiv preprint arXiv:2503.19786},
  year={2025}
}

@article{team2023gemini,
  title={Gemini: a family of highly capable multimodal models},
  author={Team, Gemini and Anil, Rohan and Borgeaud, Sebastian and Alayrac, Jean-Baptiste and Yu, Jiahui and Soricut, Radu and Schalkwyk, Johan and Dai, Andrew M and Hauth, Anja and Millican, Katie and others},
  journal={arXiv preprint arXiv:2312.11805},
  year={2023}
}

@article{an2025llava,
  title={Llava-onevision-1.5: Fully open framework for democratized multimodal training},
  author={An, Xiang and Xie, Yin and Yang, Kaicheng and Zhang, Wenkang and Zhao, Xiuwei and Cheng, Zheng and Wang, Yirui and Xu, Songcen and Chen, Changrui and Zhu, Didi and others},
  journal={arXiv preprint arXiv:2509.23661},
  year={2025}
}

@article{plizzari2024outlook,
  title={An outlook into the future of egocentric vision},
  author={Plizzari, Chiara and Goletto, Gabriele and Furnari, Antonino and Bansal, Siddhant and Ragusa, Francesco and Farinella, Giovanni Maria and Damen, Dima and Tommasi, Tatiana},
  journal={IJCV},
  year={2024}
}

@inproceedings{radford2021learning,
  title={Learning transferable visual models from natural language supervision},
  author={Radford, Alec and Kim, Jong Wook and Hallacy, Chris and Ramesh, Aditya and Goh, Gabriel and Agarwal, Sandhini and Sastry, Girish and Askell, Amanda and Mishkin, Pamela and Clark, Jack and others},
  booktitle={ICML},
  year={2021}
}

@InProceedings{Peng_2025_CVPR,
    author    = {Peng, Ruotian and He, Haiying and Wei, Yake and Wen, Yandong and Hu, Di},
    title     = {Patch Matters: Training-free Fine-grained Image Caption Enhancement via Local Perception},
    booktitle = {CVPR},
    year      = {2025}
}

@inproceedings{yin2025rod,
  title={ROD-MLLM: Towards More Reliable Object Detection in Multimodal Large Language Models},
  author={Yin, Heng and Ren, Yuqiang and Yan, Ke and Ding, Shouhong and Hao, Yongtao},
  booktitle={CVPR},
  year={2025}
}

@inproceedings{dasgupta2026,
  title={Are We There Yet? Exploring the Capabilities of MLLMs in Assistive AI Applications},
  author={Dasgupta, Shayon and Dasgupta, Avijit and Jawahar, C V},
  booktitle={ICVGIP},
  year={2025}
}

@inproceedings{lai2024lisa,
  title={Lisa: Reasoning segmentation via large language model},
  author={Lai, Xin and Tian, Zhuotao and Chen, Yukang and Li, Yanwei and Yuan, Yuhui and Liu, Shu and Jia, Jiaya},
  booktitle={CVPR},
  year={2024}
}

@inproceedings{chen2024lion,
  title={Lion: Empowering multimodal large language model with dual-level visual knowledge},
  author={Chen, Gongwei and Shen, Leyang and Shao, Rui and Deng, Xiang and Nie, Liqiang},
  booktitle={CVPR},
  year={2024}
}

@misc{openai_hello_gpt5_2025,
  author = {{OpenAI}},
  title  = {Hello {GPT}-5},
  year   = {2025},
  note   = {Blog post}
}




% LNCS Homepage, \url{http://www.springer.com/lncs}. Last accessed 4
% \end{thebibliography}
\end{document}